\documentclass{vgtc}                          

\ifpdf
  \pdfoutput=1\relax                   
  \usepackage{graphicx}                
  \DeclareGraphicsExtensions{.pdf,.png,.jpg,.jpeg} 
\else
  \usepackage{graphicx}                
  \DeclareGraphicsExtensions{.eps}     
\fi%

\graphicspath{{figures/}{pictures/}{images/}{./}} 

\usepackage{microtype}                 
\PassOptionsToPackage{warn}{textcomp}  
\usepackage{textcomp}                  
\usepackage{mathptmx}                  
\usepackage{times}                     
\usepackage{cite}                      
\usepackage{tabu}                      
\usepackage{booktabs}   
\usepackage{xcolor}
\usepackage{amsmath}
\usepackage{amssymb}
\usepackage{algorithm}
\usepackage{algpseudocode}
\usepackage{multirow}
\usepackage[normalem]{ulem}
\useunder{\uline}{\ul}{}
\onlineid{1804}

\vgtccategory{Research}

\vgtcinsertpkg

\title{Cross-View Visual Geo-Localization for Outdoor Augmented Reality}

\author{Niluthpol Chowdhury Mithun\thanks{e-mail: niluthpol.mithun@sri.com} %
\and Kshitij S. Minhas\thanks{e-mail: kshitij.minhas@sri.com} %
\and Han-Pang Chiu\thanks{e-mail: han-pang.chiu@sri.com} %
\and Taragay Oskiper\thanks{e-mail: taragay.oskiper@sri.com} %
\and Mikhail Sizintsev\thanks{e-mail: mikhail.sizintsev@sri.com} %
\and Supun Samarasekera\thanks{e-mail: supun.samarasekera@sri.com} %
\and Rakesh Kumar\thanks{e-mail:  rakesh.kumar@sri.com}
}
\affiliation{Center for Vision Technologies, SRI International, Princeton, NJ}

\abstract{ 

Precise estimation of global orientation and location is critical to ensure a compelling outdoor Augmented Reality (AR) experience. We address the problem of geo-pose estimation by cross-view matching of query ground images to a geo-referenced aerial satellite image database. Recently, neural network-based methods have shown state-of-the-art performance in cross-view matching. However, most of the prior works focus only on location estimation, ignoring orientation, which cannot meet the requirements in outdoor AR applications. We propose a new transformer neural network-based model and a modified triplet ranking loss for joint location and orientation estimation. Experiments on several benchmark cross-view geo-localization datasets show that our model achieves state-of-the-art performance. Furthermore, we present an approach to extend the single image query-based geo-localization approach by utilizing temporal information from a navigation pipeline for robust continuous geo-localization. Experimentation on several large-scale real-world video sequences demonstrates that our approach enables high-precision and stable AR insertion.

} 

\CCScatlist{
  \CCScatTwelve{Human-centered computing}{Human computer interaction (HCI)}{Interaction paradigms}{Mixed / augmented reality}{}
}

\begin{document}


\firstsection{Introduction}

\maketitle


Estimating precise 3D position and 3D orientation of ground imagery and video streams in the world is crucial for outdoor augmented reality applications. The AR system is required to insert the synthetic objects or actors at the correct spots in the imaged real scene viewed by the user. Any drift or jitter on inserted objects, which are caused by inaccurate estimation of camera poses, will disturb the illusion of mixture between rendered and real world for the user. 

Geo-localization solutions for outdoor AR applications typically rely on magnetometer and GPS sensors\cite{li2017monocular, michel2018method}. GPS provides global 3D location information, while a magnetometer measures global heading. Coupled with the gravity direction measured by IMU sensor, the entire 6-DOF (degrees of freedom) geo-pose can be estimated. However, GPS accuracy degrades dramatically in urban street canyons. Magnetometer is sensitive to external disturbance (e.g., nearby metal structures). There are also GPS-based alignment methods for heading estimation, that requires the system to be moved around at a significant distance (e.g., up to 50 meters) for initialization. In many cases, these solutions are not reliable for instantaneous AR augmentation. 

Recently, there has been a lot of interest in developing techniques for geo-localization of ground imagery using different geo-referenced data sources. Most prior works consider the problem as matching queries against a pre-built database of geo-referenced ground images or video streams. However, collecting ground images over a large area is time-consuming and may not be feasible in many cases. To overcome this limitation, there are new interests in geo-localization of ground imagery against an aerial satellite image database covering the area of interest\cite{zhu2022transgeo, zhu2021vigor, shi2022accurate, shi2019spatial, yang2021cross, shi2020looking, toker2021coming, zhao2022mutual, wang2022transformer}. Due to the wide availability and dense coverage, satellite data has become a very attractive reference data source.

\begin{figure}
    \centering  
    \vspace{0.3cm}
\includegraphics[width=0.48\textwidth]{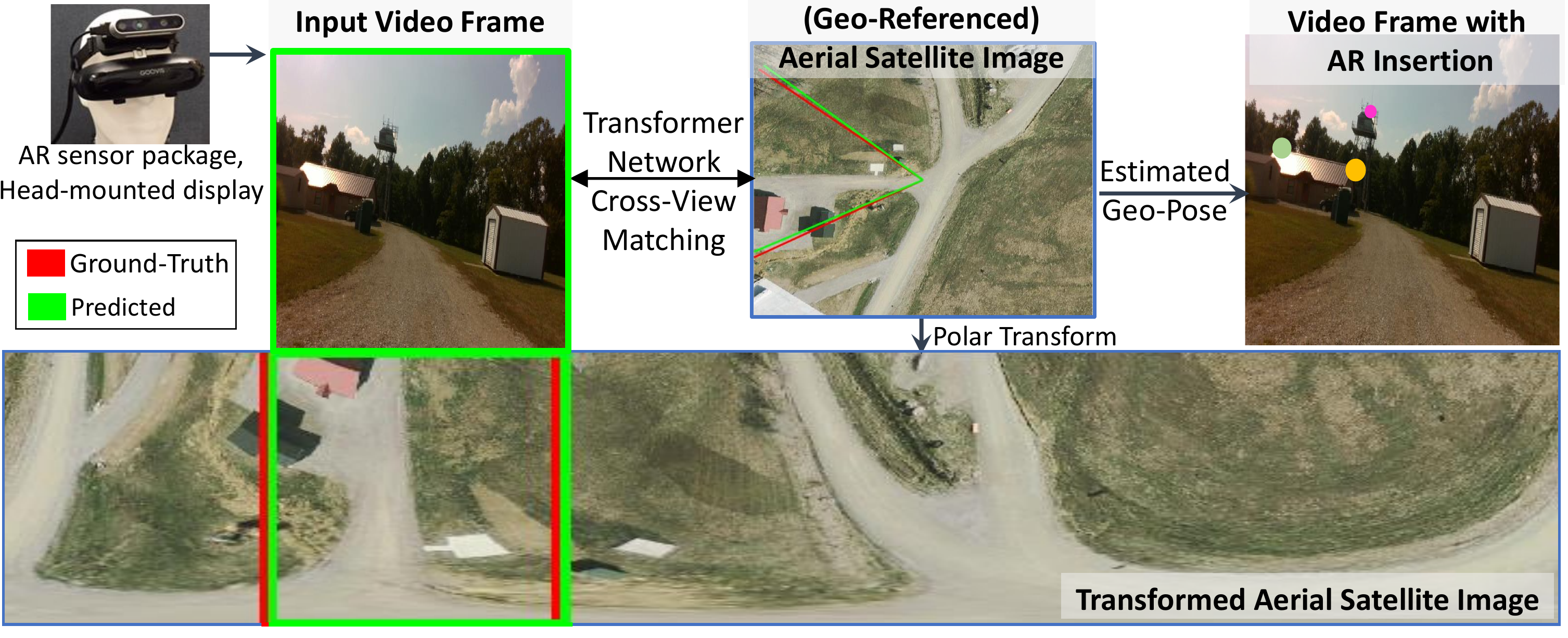}
\vspace{-0.4cm}
  \caption{Concept diagram of our proposed cross-view geo-registration approach for outdoor AR applications. A Transformer neural network-based framework matches input video frames with a geo-referenced aerial satellite image database to predict the geo-pose of the frames. We see the predicted orientation alignment (green) of the input ground image frame with respect to the polar transformed aerial satellite image, is in excellent agreement with the correct orientation (red). Based on the estimated geo-pose, our AR system can do very accurate insertions.
  }
  \vspace{-0.2cm}
  \label{fig:concept}%
\end{figure}

In this work, we present a new vision-based cross-view geo-localization solution, that matches camera images to geo-referenced aerial data sources, for outdoor AR applications (Fig.~\ref{fig:concept}). Our approach can also be used to augment existing magnetometer and GPS-based geo-localization methods. Our work focuses on matching 2D camera images to a 2D satellite reference image database, which is widely available and easier to be obtained than other 2D or 3D geo-referenced data sources. As shown in Fig.~\ref{fig:concept}, our approach enables precise AR insertions.

Specifically, we develop a navigation system to continuously estimate 6-DOF camera geo-poses for outdoor augmented reality applications. A tightly coupled visual-inertial-odometry module provides pose updates every few milliseconds. However, these pose updates will drift over time. Note the visual-inertial odometry provides robust estimates for tilt and in-plane rotation (roll and pitch) due to gravity sensing. Therefore, the drift mainly happens in the heading (yaw) and position estimates. To correct the drift, our novel cross-view visual geo-localization solution estimates 3-DOF (latitude, longitude and heading) camera pose, by matching ground camera images to aerial satellite images. This visual geo-localization solution is used both for providing initial global heading and location (cold-start procedure) and also for continuous global heading refinement over time to the navigation system. 

We propose a novel transformer neural network-based framework for our cross-view visual geo-localization solution. Compared to previous neural network models for cross-view geo-localization, our framework addresses several key limitations. First, previous works\cite{toker2021coming, zhao2022mutual, wang2022transformer, zhu2021vigor} focus only on location retrieval (considering the orientation of the query is known) utilizing the global transformer feature vector. Joint location and orientation estimation requires a spatially-aware feature representation and thus leads to a step change in our model architecture. Second, previous works use standard triplet ranking loss function, which focuses mainly on location retrieval with implicit orientation alignment\cite{yang2021cross, zhu2022transgeo}, for training the network. Instead, we modify the commonly used triplet ranking loss function to provide explicit orientation guidance. This new loss function leads to a highly accurate orientation estimation, and also helps to jointly improve location estimation accuracy. Third, previous approaches mostly assume that a ground based image panorama can be constructed at each location for query and consider each query independently, which cannot provide reliable geo-poses in a consistent manner. In contrast, we present a new approach that supports any camera movement (no panorama requirements) and utilizes temporal information from the navigation system for enabling stable and smooth AR augmentation.

\textbf{Contributions: }Main contributions are summarized as follows:

$\bullet$ We present a novel Transformer based framework for cross-view geo-localization of ground query images, by matching ground images to geo-referenced aerial satellite images. 

$\bullet$ We propose a weighted triplet loss to train our model, that provides explicit orientation guidance for location retrieval. This leads to high granularity orientation estimation and improved location estimation performance. 

$\bullet$ We achieve state-of-the-art performance in several benchmark geo-localization datasets (i.e., CVUSA\cite{zhai2017predicting}, CVACT~\cite{liu2019lending}).

$\bullet$ We extend the image-based geo-localization approach by utilizing temporal information across video frames for continuous and consistent geo-localization, which fits the demanding requirements in real-time outdoor AR applications. Experiments on several large-scale sequences demonstrate the robustness of our  cross-view visual geo-localization solution to AR applications.

\section{Related Works}

\textbf{Geo-Localization for Outdoor AR Applications:} Many approaches have been developed for registering a mobile camera in an indoor AR environment. Vision-based SLAM approaches~\cite{mur2015orb} perform quite well in such a situation. These methods can be augmented with pre-defined fiducial markers or IMU devices~\cite{li2017monocular, oskiper2012multi, oskiper2017poster} to provide metric measurements. However, they are only able to provide pose estimation in a local coordinate system, which is not suitable for outdoor AR applications. 

To make such a system work in the outdoor setting, GPS and magnetometer can be used to provide a global location and heading measurements respectively. However, the accuracy of consumer-grade GPS systems~\cite{michel2018method}, specifically in urban canyon environments, is not sufficient for many outdoor AR applications. Magnetometer also suffers from external disturbance in outdoor environments \cite{wang2020external}. 

Recently, vision-based geo-localization solutions have become a good alternative for registering a mobile camera in the world, by matching the ground image to a pre-built geo-referenced 2D or 3D database. For example, Liu et al.~\cite{liu2021net} propose a virtual-real registration approach for large-scale outdoor AR applications registering ground images to a 3D point cloud generated by aerial UAV images. Rao et al.~\cite{rao2017mobile} propose a deep learning based approach that detects objects and uses their spatial relationships for geo-registration. However, these systems completely rely on GPS and Magnetometer measurements for initial estimates. In contrast, the initial pose estimation in our system can be purely achieved by matching a ground image directly to a geo-referenced 2D satellite image database.  

There is also work~\cite{liu2018instant} that uses a panorama as input for geo-localization within a pre-built 2.5D map, which includes building footprints and heights. This system relies on segmenting the building outlines from the input image to match against the 2.5D map. It only works for specific outdoor environments. Our system uses widely available satellite data, and works in both urban and rural settings.  


\noindent \textbf{Vision-based Cross-View Geo-Localization:} Methods for visual geo-localization of ground imagery can be categorized based on geo-referenced data sources, including ground imagery collected in the past (cross-time geo-localization \cite{anoosheh2019night, seymour2019semantically, mithun2018learning}), aerial reference imagery (cross-view geo-localization \cite{workman2015wide, cai2019ground}), and data from different sensing modalities (such as LiDAR \cite{mithun2020rgb2lidar}).  

Matching ground images against satellite data has become very attractive as a visual geo-localization solution, due to the wide availability and coverage of satellite imagery. There are several benchmark datasets for evaluating these cross-view geo-localization solutions. For example, CVUSA~\cite{zhai2017predicting} and CVACT~\cite{liu2019lending} datasets provide street-view images and location-paired satellite images. 
Most cross-view geo-localization works focus on estimating the location of a given ground query. This is done by finding the closest match of this ground image to a dataset of geo-tagged satellite images. Since the viewpoint between the ground and satellite image is very different and no simple geometric transformation can be found, many methods bridge the domain gap by using polar transformations on the aerial image \cite{shi2019spatial, yang2021cross, shi2020looking}. Some methods use generative models~\cite{goodfellow2020generative} on top of the polar transforms to create more realistic images~\cite{toker2021coming, regmi2019bridging}. 

Workman et. al.~\cite{workman2015wide,workman2015location} introduced the idea of using Convolutional Neural Networks (CNNs), that were pre-trained on image classification datasets, for the task of cross-view geo-localization. Lin et al.~\cite{lin2015learning} showed significant advancements by using separate CNN branches for ground and satellite images, training with a triplet ranking loss function that brings the feature representations of matching cross-view pairs closer. Other than CNNs, transformer-based cross-view geo-localization~\cite{yang2021cross, zhu2022transgeo, wang2022transformer} methods have recently gained popularity due to their global contextual reasoning and positional dependent encoding. L2LTR~\cite{yang2021cross} first introduced transformers for cross-view localization task by implementing a hybrid ResNet~\cite{he2016deep} + ViT~\cite{dosovitskiy2020image} backbone. Their method significantly improved upon state-of-art CNN results and showed that the transformer network is well suited to the context and position-aware nature of the cross-view geo-localization task. More recently, Zhu et al.~\cite{zhu2022transgeo} proposed a two-step transformer-based method that crops out uninformative patches from the higher-resolution satellite image and showed good performance in location estimation. 

The main assumption of all these methods is that the orientation of the ground image queries is known, which is not valid for many outdoor augmented reality applications. There are very limited prior works ~\cite{shi2020looking} to consider both location and orientation estimation matching at aerial reference satellite images. For example, Shi et al.~\cite{shi2020looking} propose a CNN based joint orientation and location estimation pipeline matching to polar-transformed satellite reference images. However, they focus primarily on location retrieval with implicit orientation alignment as a by-product to help location estimation ~\cite{shi2020looking}. Moreover, the granularity of their orientation prediction is quite low. Recently, Shi et al.~\cite{shi2022beyond} propose to use a differentiable Levenberg-Marquardt optimizer to iteratively estimate the camera pose matching ground-view image features to satellite image features (projected to the ground-level viewpoint using homography with flat-ground assumption). However, \cite{shi2022beyond} relies strongly on having a good initial camera pose estimate to converge.

To the best of our knowledge, our cross-view visual geo-localization solution is the first transformer-based method that can jointly estimate the location and orientation of a ground query image matching to a geo-referenced satellite image database. In addition, we specifically extend our solution for fulfilling the demanding requirements of outdoor augmented reality applications.   








\section{Cross-View Geo-Localization Approach}

In this section, we present our cross-view geo-localization approach for finding the location and orientation of query ground images, by matching them against a geo-tagged reference satellite image database\footnote{Please note that, in the next sections, by orientation, we refer to heading, and by location, we refer to (latitude and longitude). We consider visual-inertial odometry with IMU gravity sensing can provide robust estimates for roll and pitch. Digital Elevation Model (DEM) can be used for providing estimates for height.}. Following prior works in cross-view geo-localization, we assume location-paired examples of ground images and aerial satellite images are provided for training the network. First, we present our approach for obtaining spatially-aware feature representation with a vision transformer based backbone and the dual-branch network architecture to handle cross-view pairs. After that, we discuss the details of performing orientation alignment between the ground query and polar-transformed satellite reference image. Next, we describe our proposed ranking loss function that focuses on joint orientation and location estimation. 


\begin{figure}
    \centering
  \includegraphics[width=0.48\textwidth]{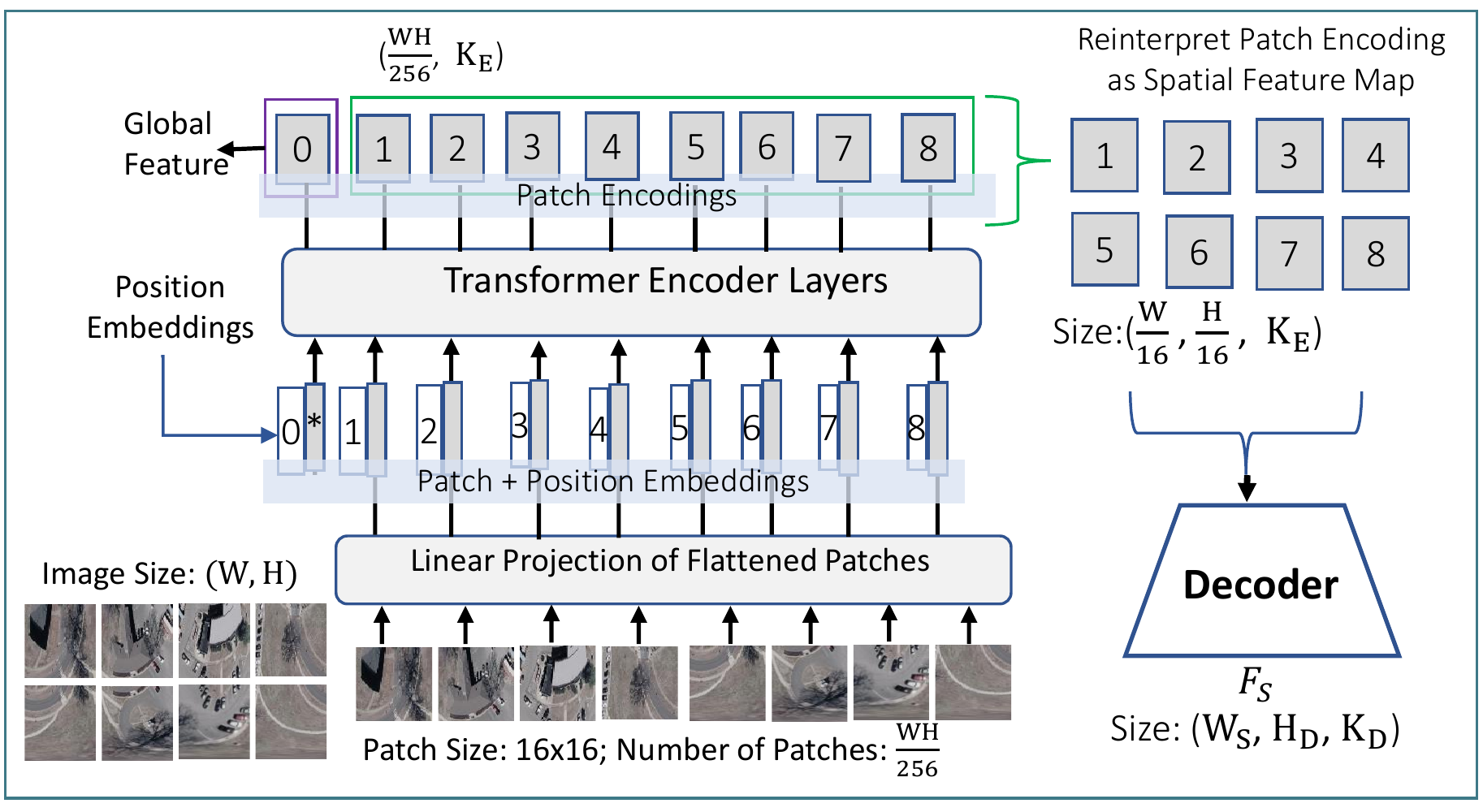}
  \caption{Illustration of our Vision Transformer based Backbone}
  \label{fig:transformer}%
\end{figure}

\subsection{Network Structure and Feature Representation}

\textbf{Transformer-based Feature Representation:} Inspired by the recent success of Vision Transformer (ViT) models in visual recognition \cite{dosovitskiy2020image}, we adopt ViT based backbone of our neural network architecture (Fig.~\ref{fig:transformer}). Vision Transformers first split an image into a sequence of fixed-size (i.e. $16 \times 16$) patches. Then the patches are flattened and embedded into tokens using a trainable linear projection layer. An extra classification token (\texttt{CLS}) is added to the sequence of embedded tokens. Position embeddings are added to the tokens to preserve position information which is crucial for vision applications. The resulting sequence of tokens are passed
through stacked transformer encoder layers. The Transformer encoder contains a sequence of blocks consisting of multi-headed self-attention module and a feed-forward network. The feature encoding corresponding to \texttt{CLS} token is considered as global feature representation and  prior works on transformer-based models for cross-view geo-localization \cite{yang2021cross} uses the global feature vector considering it as a pure location estimation problem. However, to address the problem of joint location and orientation estimation, it is crucial to extract spatial-aware feature representation that retains local positional information. 

To address this we design a simple up-sampling decoder following the transformer encoder. The decoder alternates convolutional layers and bilinear upsampling operations (with upsampling restricted to $2x$). Based on the patch features from the transformer encoder, the decoder is used to attain the target spatial feature resolution. As the goal of encoder-decoder model is to generate a spatial-aware representation, we first reshape the sequence of patch encoding from a 2D shape of size $\frac{WH}{256} \times K_E$ to a 3D feature map of size $\frac{W}{16} \times \frac{H}{16} \times K_E$. Here, ($W$, $H$) is the resolution of the input image and $K_E$ is the patch embedding vector size. The decoder takes this 3D feature map as input and outputs the final spatial feature representation $\mathbb{F}$.

\begin{figure}[t]
    \centering    \includegraphics[width=0.48\textwidth]{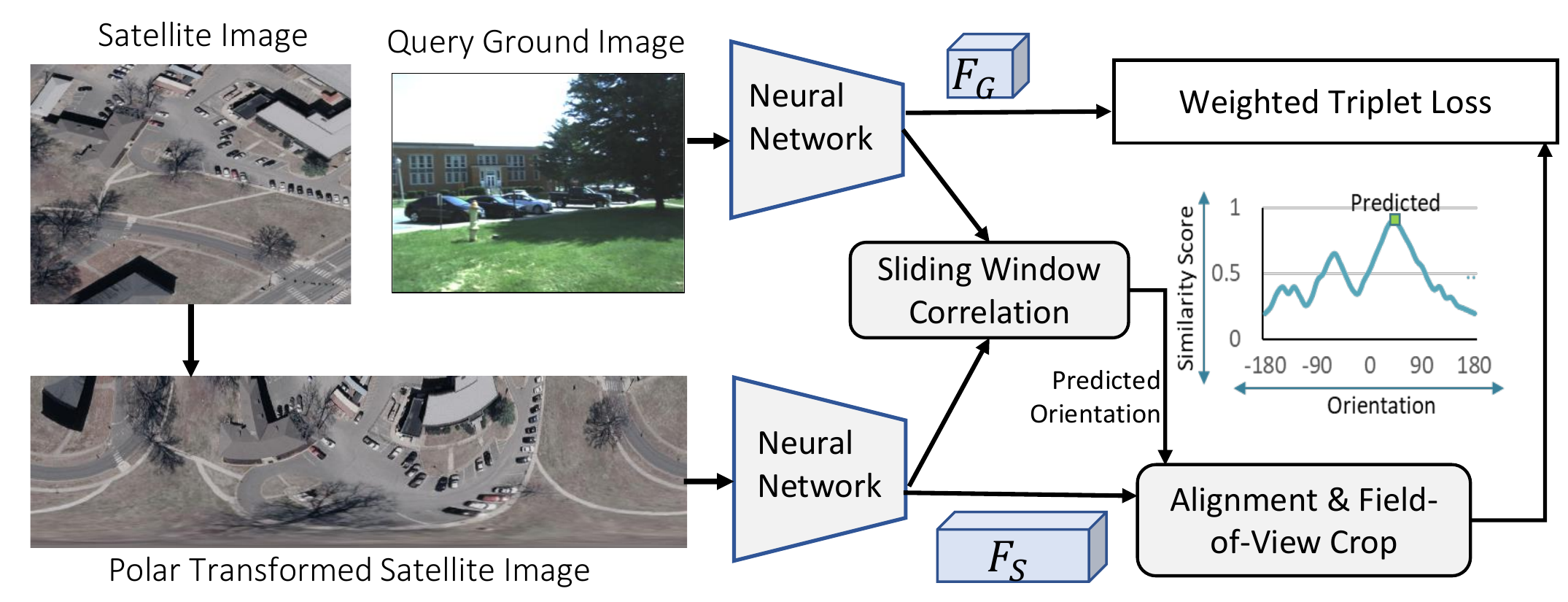}
  \caption{Brief Illustration of Cross-View Geo-Localization Approach}
  \label{fig:framework}%
\end{figure}

\vspace{0.1cm}
\noindent \textbf{Network Architecture:} We use a two-branch neural network architecture to train our model using location-coupled pairs of ground images and aerial satellite images (Fig.~\ref{fig:framework}). One of the branches focuses on encoding ground images and the other branch focuses on encoding aerial reference images. Both of the branches consist of a Transformer-based encoder-decoder backbone as described above. 





\subsection{Cross-View Orientation Alignment} 

\textbf{Polar Transformation of Reference Satellite Image:} Due to drastic viewpoint changes between cross-view ground and aerial images, we follow prior work \cite{shi2019spatial, shi2020looking} to apply the polar transformation on satellite images that focuses on projecting satellite image pixels to the ground-level coordinate system. Polar transformed satellite images are coarsely geometrically aligned with ground images and it is used as a pre-processing step to reduce the cross-view spatial layout gap. The width of the polar transformed image is constrained to be proportional to the field of view in the same measure as the ground images. Hence, when the ground image has an FoV of $360$ degrees(i.e., panorama), the width of the ground image should be of the same size as the polar transformed image. Additionally, the polar transformed image is constrained to have the same vertical size (i.e., height) as the ground images.

\vspace{0.1cm}
\noindent \textbf{Orientation Alignment by Sliding Window Matching:} Now, we describe the sliding window-based similarity matching approach to estimate orientation alignment between ground query and aerial reference. Estimating the orientation of the ground query is very important and it also helps in more precise location estimation. The orientation alignment between cross-view images is estimated based on the assumption that the feature representation of the ground image and polar transformed aerial reference image should be very similar when they are aligned. In this regard, we slide a search window (of the size of the ground feature) along the horizontal direction (i.e., orientation axis) of the feature representation obtained from the aerial image and compute the similarity of the ground feature with the aerial reference features at all the possible orientations. The horizontal position corresponding to the highest similarity is considered to be the orientation estimate of the ground query with respect to the polar-transformed aerial one.

Let's consider we have a ground-satellite image pair $(I_G, I_S)$. The spatial feature representation for this pair is denoted as $(\mathbb{F}_G, \mathbb{F}_S)$. Here, $\mathbb{F}_S\in R^{W_S \times H_D \times K_D}$ and $\mathbb{F}_G\in R^{W_G \times H_D \times K_D}$. Here, ($W_S$, $H_D$) is the resolution of the satellite image feature representation, and ($W_G$, $H_D$) is the resolution of the ground image feature representation. $K_D$ is the number of channels. When the ground image is a panorama, $W_G$ is the same as $W_S$. Otherwise, $W_G$ is smaller than $W_S$. The similarity $\mathbb{S}$ between $\mathbb{F}_G$ and $\mathbb{F}_S$ at the horizontal position $i$ can be calculated as follows:

\begin{equation}
    \mathbb{S}(i) = \sum_{k=1}^{K_D} \sum_{h=1}^{H_D} \sum_{w=1}^{W_G}  \mathbb{F}_G[w,h,k] \ \mathbb{F}_S[(w\!+\!i)\!\mathbin{\%}\!W_S,h,k],
    \label{eqn:ori_sim}
\end{equation}
where $\mathbin{\%}$ denotes modulo operator. $\mathbb{F}[w,h,k]$ denotes the feature activation at index $(w, h, k)$ and $i=\{1,...W_S\}$. The granularity of orientation prediction depends on the size of $W_S$. As there are $W_S$ possible orientation estimates and hence, orientation prediction is possible for every $\frac{360}{W_S}$ degree. Hence, a larger size of $W_S$ would allow orientation estimation at a finer scale. From the calculated similarity vector $\mathbb{S}$, the position of the maximum value of $\mathbb{S}$ is the estimated orientation of the ground query. Let, $\mathbb{S}_{Max}$ denote the maximum value of similarity scores and $\mathbb{S}_{GT}$ is the value of similarity score at the ground-truth orientation. When $\mathbb{S}_{Max}$ and $\mathbb{S}_{GT}$ are the same, we have perfect orientation alignment between query and reference images.

\subsection{Training Objective}

After the feature embeddings from the ground and satellite image pairs are extracted, the goal is to learn to maximize the similarity between the feature embeddings of ground images and corresponding paired satellite images. 
Prior work \cite{shi2020looking} uses the typical soft-margin triplet loss and expects that network will implicitly perform coarse orientation alignment while learning to perform location retrieval. However, due to no explicit orientation guidance in the training objective, this approach is only suitable for coarse orientation alignment to help location search. To enforce the model to learn precise orientation alignment and location estimation jointly, we propose an orientation-weighted triplet ranking loss.
\begin{equation}
    \mathcal{L}_{T} = \mathcal{W}_{Ori}  *\mathcal{L}_{GS}.
    \label{eqn:total_loss}
\end{equation}

Here, $\mathcal{L}_{GS}$ is a soft margin triplet ranking loss that attempts to bring feature embeddings of matching pairs closer while pushing the feature embeddings of
not matching pairs far apart. $\mathcal{L}_{GS}$ is defined as follows,
\begin{equation}
    \mathcal{L}_{GS} = \log\big(1+e^{\alpha(||\mathbb{F}_G-\mathbb{F}_{S}||_F - ||\mathbb{F}_G-\mathbb{F}_{\hat{S}}||_F)}\big),
\end{equation}
where $\mathbb{F}_{\hat{S}}$ is a non-matching satellite image feature embedding for ground image feature embedding $\mathbb{F}_G$, and $\mathbb{F}_S$ is the matching (i.e., location paired) satellite image feature embedding. $||.||_F$ denotes the Frobenius norm. Parameter $\alpha$ is used to adjust the convergence speed of training. The loss term attempts to ensure that for each query ground image feature, the distance with the matching cross-view satellite image feature is smaller than the distance with the non-matching satellite image features.

We weight the triplet ranking loss function based on the orientation alignment accuracy with the weighting factor $\mathcal{W}_{Ori}$. It attempts to provide explicit guidance based on orientation alignment similarity scores (Eqn.~\ref{eqn:ori_sim}),
which is defined as follows:
\begin{equation}
    \mathcal{W}_{Ori} = 1 + \beta * \frac{\mathbb{S}_{Max} - \mathbb{S}_{GT}}{\mathbb{S}_{Max} - \mathbb{S}_{Min}},
    \label{eqn:ori_wt}
\end{equation}
where $\beta$ is a scaling factor.  $\mathbb{S}_{Max}$ and $\mathbb{S}_{Min}$ are the maximum and minimum value of similarity scores respectively. $\mathbb{S}_{GT}$ is the similarity score at the ground-truth position. The weighting factor $\mathcal{W}_{Ori}$ attempts to apply a penalty on the loss when $\mathbb{S}_{Max}$ and $\mathbb{S}_{GT}$ are not the same.

\section{Outdoor AR System with Geo-Localization}

\begin{figure}[t]
    \centering
    \includegraphics[width=0.46\textwidth]{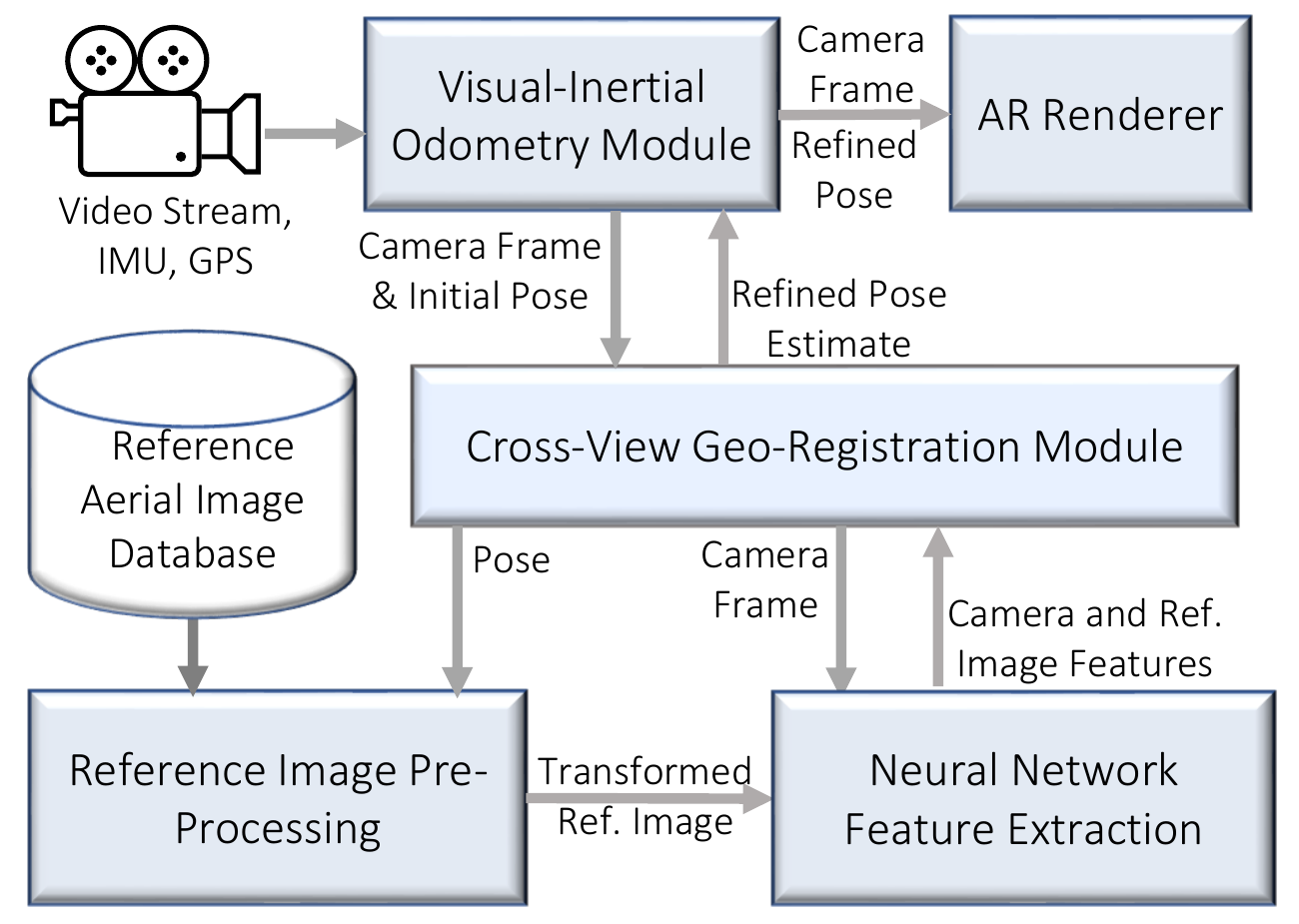}
  \caption{AR System with Geo-Localization Pipeline}
  \label{fig:system}%
\end{figure}

In this section, we discuss our outdoor AR system with cross-view geo-localization (Fig.~\ref{fig:system}). First, we briefly describe our Visual-Inertial Odometry module. Then, we discuss the reference image pre-processing module. Finally, we present our Geo-Registration module which enables continuous geo-localization using a sequence of images. The Neural Network Feature Extraction module is based on the trained neural network model described in Sec.3.

\subsection{Navigation Pipeline}
For real-time 6-DOF geo-pose estimation, the navigation (Nav.) pipeline in our AR system uses a sensor package that includes hardware synchronized inertial measurement unit (IMU), a stereo pair of cameras, RGB color camera, and a GPS device. Raw sensor readings from both the IMU and the stereo cameras are fed into the navigation pipeline for tracking. A high-resolution RGB camera is used for AR augmentation. The two videos from the stereo pair of cameras are used by the navigation pipeline for tracking. These stereo frames are not used by the cross-view geo-registration module, which uses the RGB camera frames. For video see-through (VST) AR, the module is able to produce poses at the RGB camera frame rate with minimal latency. 

The Visual-Inertial Odometry (VIO) module is implemented based on prior works~\cite{mourikis2007multi,oskiper2012multi}. For visual-inertial navigation, a tightly coupled error-state Extended Kalman Filter (EKF) based sensor fusion architecture is utilized. In addition to the relative measurements from frame-to-frame feature tracks for odometry purposes, the error-state EKF framework is capable of fusing global measurements from GPS and refined estimates from the Geo-Registration module, for heading and location correction to counter visual odometry drift accumulation over time.

The EKF state vector consists of errors in location, orientation, and velocity together with errors in gyro and accelerometer biases~\cite{oskiper2012multi}. In the case of orientation correction, the error measurements are constructed as the difference between the global measurement returned from the cross-view geo-registration process and the predicted value from the EKF filter. For details please confer~\cite{oskiper2012multi, oskiper2015augmented}, which describes the global orientation correction from a geo-landmarking process where the orientation measurements are constructed by manually establishing a 3D to 2D correspondence between a known world landmark feature and its manually clicked position in the video image. In this work, the same fusion mechanism is applied to the orientation information returned by the automatic Geo-Registration pipeline.

\begin{algorithm}[t]
\caption{Global Orientation Estimation of Image Sequences}\label{algo:seq}
\begin{algorithmic}[]
\State
{\bf Input:} Continuous Streaming Video and Pose from Navigation Pipeline.

\State
{\bf Output:} Global orientation estimates, $\{ q_t|t=0,1,... \}$
\State
{\bf Parameters:} The maximum length of frame sequence used for orientation estimation $\tau$. FoV coverage threshold $\delta_F$. Ratio-test threshold $\delta_R$.

\State
{\bf Initialization:} Initialize dummy orientation $y_{0}$ of the first Camera Frame $V_{0}$ to zero.

\vspace{0.07cm}
\State
\quad{\bf Step 0:} Learn the two-branch cross-view geo-localization neural network model (Sec.3) using the training data available.

\State
\quad{\bf Step 1:} Receive Camera Frame $V_t$, Camera global position estimate $G_t$ and Relative Local Pose $P_t$ from Navigation Pipeline at time step $t$. 

\State
\quad{\bf Step 2:} Calculate the relative orientation between frame $t$ and $t-1$ using local pose $P_t$. This relative orientation is added to the dummy orientation at frame $t-1$ to calculate $y_t$.  $y_t$ is used to track orientation change with respect to the first frame.

\State
\quad{\bf Step 3:} Collect an aerial satellite image centered at position $G_t$. Perform polar-transformation on the image.

\State
\quad{\bf Step 4:} Apply the trained two-branch  model to extract feature descriptors $F_G$ and $F_S$ of camera frame and polar-transformed aerial reference image respectively.

\State
\quad{\bf Step 5:} Compute the similarity $S_t$ of the ground image feature with the aerial reference feature at all possible orientations using the ground feature as a sliding window. 

\State
\quad{\bf Step 6:}  Put ($S_t$, $y_{t}$) in Buffer $\mathrm{B}$. If the Buffer $\mathrm{B}$ contains more than $\tau$ samples, remove the sample ($S_{t-\tau}$, $y_{t-\tau}$) from Buffer.

\State
\quad{\bf Step 7:} Using Buffer $\mathrm{B}$, accumulate orientation prediction score over frames into $S_{t}^{\mathbf{A}}$. Before accumulation, the similarity score vectors for all the previous frames are circularly shifted based on the difference in their respective dummy orientations with $y_t$. The position corresponding to the highest similarity in $S_{t}^{\mathbf{A}}$ is the orientation estimate based on the frame sequence in $\mathrm{B}$. 

\State
\quad{\bf Step 8:} Calculate FoV coverage of the frames in Buffer using dummy orientations. Find all the local maxima in the accumulated similarity score $S_{t}^{\mathbf{A}}$. Perform ratio test based on the best and second-best maxima score. 

\State
\quad{\bf Step 9:} If FoV coverage and lowe's ratio text score are more than $\delta_F$ and $\delta_R$ respectively, the estimated orientation measurement $q_t$ is selected and sent to be used by the navigation module to refine pose estimate. Otherwise, inform the navigation module that the estimated orientation is not reliable. 

\State
\quad{\bf Step 10:} Go to Step 1 to get the next set of frame and pose from the navigation pipeline.  

\end{algorithmic}
\end{algorithm}

\subsection{Reference Image Pre-Processing Module}
When training and testing with benchmark data-sets, cropped aerial satellite images are already available. In case of utilizing the cross-view geo-localization approach with AR system, we consider having access to a geo-referenced Satellite/Ortho image completely covering the region of our interest. Given the geo-location (i.e., Latitude, Longitude) of the camera frame, we crop a square from the satellite image centered at the camera location. As the ground resolution of satellite images varies across areas, we ensure the image covers an approximately same-size area as in the training dataset (e.g., the aerial images in CVACT~\cite{liu2019lending} dataset cover approximately $144m \times 144m$ area). Hence, the size of the aerial image crop depends on the ground resolution for satellite images. This aerial image is then transformed into a polar-transformed image of size the same as that of the ground image with 360-degree Field of View.

\subsection{Cross-View Geo-Registration Module}

For a single camera frame, the highest similarity score along the horizontal direction matching the satellite reference usually serves as a good orientation estimate. However, a single frame might have quite limited context especially when the camera FoV is small. Hence, there is a possibility of significant ambiguity in many cases and the estimate is unlikely to be reliable/stable for outdoor AR. Moreover, we have access to frames continuously from the navigation pipeline and it is desirable to jointly consider multiple sequential frames to provide a high-confidence and stable estimate.  In this regard, we extend the single image-based cross-view matching approach (as described in the previous section) to using a continuous stream of images and relative poses between them from the Nav. Pipeline.

When the navigation module is equipped with a GPS, we only need to perform orientation estimation. The algorithm for global orientation estimation is summarized in Algo.~\ref{algo:seq}. Our approach can be used for both providing a cold-start geo-registration estimate at the start of the AR system and also for refining the estimates continuously after the cold-start is complete. In the case of continuous refinement, we consider a smaller search range (i.e., $\pm 6$ degrees for orientation refinement) around the initial estimate from the navigation pipeline. Please note that at the time of cold-start, we expect a user to rotate his/her head, standing in one place  so that the frame sequence has sufficient FoV coverage for a reliable prediction. We perform outlier removal based on FoV coverage of the frame sequence and lowe's ratio test \cite{lowe2004distinctive} (comparing the best and the second best local maxima in the accumulated similarity score). A larger value of FoV coverage and ratio test indicates high confidence prediction. In the case of continuous refinement, only the ratio test score is used for outlier removal.


In a GPS-challenged scenario, we need to estimate both location and orientation. We assume to have a crude estimate of location and we select a search region based on location uncertainty (e.g., $100m \times 100m$). In the search region, we sample locations every  $x_s$ meters (e.g., $x_s$=$2$). For all the sampled locations, we create a reference image database collecting satellite image crop centered at the location. Next, we calculate the similarity between the camera frame at time $t$ and all the reference images in the database. After the similarity calculation, we select the top $N$ (e.g., $N=25$) possible matches based on the similarity score to limit the run-time of subsequent estimation steps. Then, these matches are verified based on whether they are consistent over a short frame sequence $f_d$ (e.g. $f_d$ = $20$). For each of the selected $N$ sample locations, we calculate the next probable reference locations using the relative pose for the succeeding sequence of frames of length $f_d$. This provides us with an $N$ set of reference image sequences of size $f_d$. If the similarity score with the camera frames is higher than the selected threshold for all the reference images in a sequence, the corresponding location is considered consistent. If this approach returns more than one consistent result, we select the one with the highest combined similarity score. The best orientation alignment with the selected reference image sequence is selected as the estimated orientation. Please note that we expect a user to rotate his/her head, standing in one place at the cold-start stage. In such a case of the user standing in one place, each of the $N$ sets will contain one reference image as there is no position displacement. This allows significant speed-up as we do not need to compute reference image feature descriptors again after computing only once at the start. 

\section{Experiments}

In this section, we first present results on several benchmark datasets validating our cross-view geo-localization approach. Then, we evaluate the robustness and applicability of our outdoor AR system with cross-view geo-localization module on several real-world navigation sequences collected across the United States.

\subsection{Experiments on Benchmark Cross-View Geo-Localization Datasets}

We first describe the datasets, evaluation metrics, and 
implementation details (Sec.~\ref{datasets}). Then, we evaluate the performance of our proposed approach on cross-view location estimation by comparing several state-of-the-art approaches (Sec.~\ref{quantitative_loc}). Next, we evaluate our approach on the task of cross-view orientation estimation by comparing it with state-of-the-art approaches and baselines (Sec.~\ref{quantitative_loc}).


\subsubsection{Datasets, Metrics and Training Details} 
\label{datasets}

\vspace{0.1cm}
\noindent \textbf{Datasets:} We test our pipeline on two standard benchmark cross-view localization datasets (i.e., CVUSA~\cite{zhai2017predicting}, CVACT~\cite{liu2019lending}). First, is the CVUSA\cite{zhai2017predicting} dataset which contains 35,532 ground and satellite image pairs for training and 8,884 pairs for testing. The images in the CVUSA dataset are collected across the USA, whereas the images in CVACT are collected in Australia. CVACT~\cite{liu2019lending} dataset gives us the same amount of pairs for training and testing (CVACT\_val). Both these datasets provide ground panorama images and corresponding location-paired satellite images. The ground and aerial images are north-aligned in these datasets. The CVACT dataset also provides the GPS locations along with the ground–satellite image pairs.



\vspace{0.1cm}
\noindent \textbf{Evaluation Metrics: } We evaluate our system on both cross-view location and orientation estimation tasks. For evaluation, we use standard evaluation metrics following prior cross-view image localization works \cite{shi2019spatial, shi2020looking, toker2021coming,zhu2022transgeo, yang2021cross}. For location estimation, We report results with rank-based $R@k$ (Recall at $k$) metric to compare the performance of our method with the state-of-the-art. $R@k$ calculates the percentage of queries for which the ground truth (GT) results are found within the top-$k$ retrievals (higher is better). Specifically, we find the top-$k$ closest satellite image embeddings to a given ground panorama image embedding. If the paired satellite embedding is present with top-$k$ retrieval, then it is considered a success. We report results for
$R@1$, $R@5$, and $R@10$.

For orientation estimation, we predict the orientation of
query ground images considering we know the geo-location of the queries (i.e., the paired aerial reference image is known). Orientation estimation accuracy is calculated based on the difference between predicted and GT orientation (i.e., orientation error). If the orientation error is within the threshold $j$ (in degrees), the estimated orientation estimation is deemed as correct. In experiments, we report results for $j$ $=$ $2$, $4$, $6$, $12$ degrees to compare performance with other methods.


\vspace{0.1cm}
\noindent \textbf{Training Details:} The proposed network is implemented in PyTorch. We used two NVIDIA GTX 1080Ti GPUs to train our models. We use 128 x 512-sized ground panorama images, and the paired satellite images are also polar-transformed to the same size. Following~\cite{yang2021cross}, we trained the models using AdamW\cite{loshchilov2018fixing} optimizer with a cosine learning rate schedule and learning rate of 1e-4. We start with ViT\cite{dosovitskiy2020image} model pre-trained on ImageNet\cite{deng2009imagenet} dataset and train for 100 epochs with a batch size of 16.

\subsubsection{Location Estimation}
\label{quantitative_loc}
We report the performance of our approach on location estimation on CVUSA dataset in Table~\ref{table:loc_cvusa} and on CVACT dataset in Table~\ref{table:loc_cvact}. We compare our approach with several state-of-the-art cross-view location retrieval approaches, i.e., SAFA \cite{shi2019spatial}, DSM~\cite{shi2020looking}, Toker et al.\cite{toker2021coming}, L2LTR~\cite{yang2021cross}, TransGeo \cite{zhu2022transgeo}, TransGCNN~\cite{zhao2022mutual}, MGTL \cite{wang2022transformer}. For these compared approaches, we directly cite the best-reported results from respective papers when available. Among these compared approaches, SAFA \cite{shi2019spatial}, DSM~\cite{shi2020looking}, Toker et al.\cite{toker2021coming} use CNN-based backbones, whereas the other approaches including our proposed approach use Transformer based backbones.

It is evident from Table~\ref{table:loc_cvusa} and Table~\ref{table:loc_cvact} that our method performs better than other methods in all the evaluation metrics. We observe that all the transformer-based approaches achieve large performance improvement over CNN-based approaches. For example, the best CNN-based method (i.e., Toker et al.\cite{toker2021coming}) achieves $R@1$ of 92.56 in CVUSA and 83.28 in CVACT, whereas the best Transformer-based approach (i.e., ours proposed) achieves significantly higher $R@1$ of 94.89 in CVUSA and $85.99$ in CVACT. This is expected due to the great recent success of Transformer models in  vision applications. Among the Transformer-based approaches, the proposed approach works the best. We believe the joint location and orientation estimation capability of our proposed approach helps to better handle the cross-view domain gap.




\renewcommand{\arraystretch}{1.1}
\begin{table}[]
\centering
\caption{Comparisons of location estimation results with state-of-the-art methods on CVUSA\cite{zhai2017predicting} dataset.}
\begin{tabular}{l|l|l|l}
\hline  
\multirow{2}{*}{Method}   & \multicolumn{3}{c}{\underline{Evaluation Metrics}}       \\ 
         & R@1     & R@5   & R@10  \\
         \hline  
SAFA \cite{shi2019spatial}          & 89.84 & 96.93 & 98.14 \\
DSM \cite{shi2020looking}           & 91.96 & 97.50 & 98.54 \\
Toker et al.\cite{toker2021coming}  & 92.56 & 97.55 & 98.33 \\
L2LTR \cite{yang2021cross}          & 94.05 & 98.27 & 98.99 \\
TransGeo \cite{zhu2022transgeo}     & 94.08 & 98.36 & 99.04 \\
TransGCNN \cite{zhao2022mutual}     & 94.12 & 98.21 & 98.94 \\
MGTL \cite{wang2022transformer}     & 94.11 & 98.30 & 99.03 \\
\textbf{Proposed} & \textbf{94.89}   & \textbf{98.65} & \textbf{99.33} \\
\hline
\end{tabular}
\label{table:loc_cvusa}
\end{table}
\renewcommand{\arraystretch}{1}

\renewcommand{\arraystretch}{1.1}
\begin{table}[]
\centering
\caption{Comparisons of location estimation results with state-of-the-art methods on CVACT~\cite{liu2019lending} dataset.}
\begin{tabular}{l|l|l|l}
\hline  
\multirow{2}{*}{Method}   & \multicolumn{3}{c}{\underline{Evaluation Metrics}}       \\ 
         & R@1     & R@5   & R@10  \\
         \hline  
SAFA \cite{shi2019spatial}          & 81.03 & 92.80 & 94.84 \\
DSM \cite{shi2020looking}           & 82.49	& 92.44	& 93.99 \\
Toker et al.\cite{toker2021coming}  & 83.28 & 93.57 & 95.42 \\
L2LTR \cite{yang2021cross}          & 84.89	& 94.59	& 95.96 \\
TransGeo \cite{zhu2022transgeo}     & 84.95 & 94.14 & 95.78 \\
TransGCNN \cite{zhao2022mutual}     & 84.92 & 94.46 & 95.88 \\
MGTL \cite{wang2022transformer}     & 85.35 & 94.45 & 95.96 \\
\textbf{Proposed} & \textbf{85.99}	& \textbf{94.77}	& \textbf{96.14} \\
\hline
\end{tabular}
\label{table:loc_cvact}
\end{table}
\renewcommand{\arraystretch}{1}

\subsubsection{Orientation Estimation}
\label{quantitative_ori}
We report the orientation estimation performance of the proposed approach comparing other approaches in Table~\ref{tab:orien_cvusa} and Table~\ref{tab:orien_cvact}. We divide Table~\ref{tab:orien_cvusa} into two rows to aid our analysis. Among the compared approaches, CNN-based DSM~\cite{shi2020looking} is the state-of-the-art in cross-view orientation estimation. We also report the performance of the ViT-based L2LTR~\cite{yang2021cross} model (trained for location estimation) for orientation estimation by reinterpreting its patch encoding as a spatial feature map. As there are no prior Transformer-based models trained for orientation estimation, we report L2LTR baseline to analyze how Transformer-based models trained on location estimation work on orientation estimation. In Table~\ref{tab:orien_cvusa} row-3.1 and Table~\ref{tab:orien_cvact}, we compare these approaches with the proposed. It is evident that the proposed approach shows huge improvements not only in orientation estimation accuracy but also in the granularity of prediction.  

To further analyze the components of our model, we implement several baselines and report in row-3.2 of Table~\ref{tab:orien_cvusa}. As the DSM~\cite{shi2020looking} network architecture only trains to estimate orientation at a granularity of 5.6 degrees compared to 1 degree in ours, a fair comparison is not directly possible. We extend DSM model by removing some pooling layers in CNN model and changing the input size so that orientation estimation at 1-degree granularity is possible. We name this baseline ``DSM-360". The second baseline in row 3.2 is ``DSM-360 w/ $\mathcal{L}_T$" which trains DSM-360 with our proposed loss.Comparing the performance of DSM-360 and DSM-360 w/ $\mathcal{L}_T$ with the proposed, we observe that our Transformer-based model shows significant performance improvement across orientation estimation metrics. For example, our proposed approach achieves orientation error with 2 Degrees (Deg.) for $93\%$ of ground image queries, whereas DSM-360 achieves $88\%$. We also observe that DSM-360 trained with the proposed $\mathcal{L}_T$ loss achieves consistent performance improvement over DSM-360. However, the performance is still significantly lower than the proposed method. The third baseline in row 3.2 is ``Proposed w/o $\mathcal{W}_{Ori}$". This baseline follows the proposed network architecture, but it is trained with standard soft-margin triplet loss $\mathcal{L}_{GS}$ (i.e., without any orientation estimation based weighting $\mathcal{W}_{Ori}$). We observe that for higher orientation error ranges (e.g., 6 deg., 12 deg.), this baseline shows comparable results to the proposed. However, for finer orientation error ranges (e.g., 2 deg.), we find a drastic drop in performance. From these results, it is evident that the proposed weighted loss function is crucial for the model to learn to handle ambiguities in fine-grained geo-orientation estimation.






\renewcommand{\arraystretch}{1.1}
\begin{table}[t]
\caption{Comparisons of orientation estimation results with state-of-the-art methods and baselines on CVUSA\cite{zhai2017predicting} dataset. We report performance for prediction range 64 (as reported in prior work~\cite{shi2020looking}) on row-\ref{tab:orien_cvusa}.1. In row-\ref{tab:orien_cvusa}.2, we present the performance of some baselines implemented by us for prediction range 360.}
\centering
    \resizebox{1\linewidth}{!}{
\begin{tabular}{c| @{\hspace{0.5\tabcolsep}} @{}p{2.6cm}@{}|@{\hspace{0.3\tabcolsep}} c @{\hspace{0.3\tabcolsep}}|@{\hspace{0.35\tabcolsep}} c @{\hspace{0.35\tabcolsep}}|@{\hspace{0.35\tabcolsep}} c @{\hspace{0.35\tabcolsep}}|@{\hspace{0.35\tabcolsep}} c @{\hspace{0.35\tabcolsep}}|@{\hspace{0.3\tabcolsep}} c @{\hspace{0.2\tabcolsep}}}
\hline
\multirow{2}{*}{\#} & \multirow{2}{*}{\parbox{1.4cm}{\centering Method}}  & \multirow{2}{*}{\parbox{1.4cm}{\centering Prediction Range}} & \multicolumn{4}{c}{\underline{Orientation Error Range}}    \\
     & &  & 2 Deg.              & 4 Deg. & 6 Deg. & 12 Deg. \\
     \hline
 & L2LTR \cite{yang2021cross} & 64               & -                       & -         & 0.27      & 0.54       \\
\ref{tab:orien_cvusa}.1 & DSM \cite{shi2020looking}   & 64               & -                       & -         & 0.85      & 0.90       \\
& DSM w/ $\mathcal{L}_{T}$  & 64               & -                       & -         & 0.89      & 0.94       \\
& \textbf{Proposed}   & 64              & -                    & -      & \textbf{0.94}      & \textbf{0.98} \\
\hline \hline
 & DSM-360  & 360     & 0.88 & 0.93         & 0.93       & 0.95       \\
\ref{tab:orien_cvusa}.2 & DSM-360 w/ $\mathcal{L}_{T}$   & 360     & 0.89 & 0.95         & 0.96       & 0.97       \\
& Proposed w/o $\mathcal{W}_{Ori}$   & 360              & 0.77	&0.93	&0.97	&0.98 \\
& \textbf{Proposed}   & 360              & \textbf{0.93}                    & \textbf{0.97}      & \textbf{0.98}      & \textbf{0.99} \\
\hline
\end{tabular}}
\label{tab:orien_cvusa}
\end{table}
\renewcommand{\arraystretch}{1}

\renewcommand{\arraystretch}{1.1}
\begin{table}[t]
\caption{Comparisons of orientation estimation results with state-of-the-art methods on CVACT~\cite{liu2019lending} dataset.}
\centering
    \resizebox{0.98\linewidth}{!}{
\begin{tabular}{l|c|c|c|c|c}
\hline
\multirow{2}{*}{Method}  & \multirow{2}{*}{\parbox{1.4cm}{\centering Prediction Range}} & \multicolumn{4}{c}{\underline{Orientation Error Range}}    \\
     &  & 2 Deg.              & 4 Deg. & 6 Deg. & 12 Deg. \\
     \hline
L2LTR \cite{yang2021cross} & 64               & -                       & -         & 0.39      & 0.66       \\
DSM \cite{shi2020looking}   & 64               & -                       & -         & 0.89       & 0.97       \\
\textbf{Proposed}    & \textbf{360}              & \textbf{0.95}                    & \textbf{0.98}      & \textbf{0.99}     & \textbf{0.99} \\
\hline
\end{tabular}}
\label{tab:orien_cvact}
\end{table}
\renewcommand{\arraystretch}{1}

\subsection{Experiments on Real-World Navigation Sequences}

\renewcommand{\arraystretch}{1.1}
\begin{table}[t]
\caption{Continuous full $360^{\circ}$ heading estimation performance on real-world navigation sequences.}
\vspace{-0.2cm}
\begin{center}
\begin{tabular}{@{\hspace{1.5\tabcolsep}} l @{\hspace{1.5\tabcolsep}}|@{\hspace{1.5\tabcolsep}} l @{\hspace{3\tabcolsep}} l @{\hspace{3\tabcolsep}} l}
\hline
FoV Coverage                  & Any  & $120^{\circ}$ & $180^{\circ}$ \\
\hline \hline
\multicolumn{4}{c}{Set 1, Mercer County, New Jersey}         \\
\hline
Accuracy ($\pm 2^{\circ}$)      & 0.60 & 0.64    & 0.69   \\
Accuracy ($\pm 5^{\circ}$)      & 0.87 & 0.88    & 0.90    \\
Accuracy ($\pm 10^{\circ}$)     & 0.96 & 0.99    & 0.99   \\
\hline \hline
\multicolumn{4}{c}{Set 2, Prince William County, Virginia}         \\
\hline
Accuracy ($\pm 2^{\circ}$)      & 0.54 & 0.56    & 0.63    \\
Accuracy ($\pm 5^{\circ}$)      & 0.81 & 0.83    & 0.90    \\
Accuracy ($\pm 10^{\circ}$)     & 0.93 & 0.94    & 0.95    \\
\hline \hline
\multicolumn{4}{c}{Set 3, Johnson County, Indiana}         \\
\hline
Accuracy ($\pm 2^{\circ}$)      & 0.61 & 0.61    & 0.71    \\
Accuracy ($\pm 5^{\circ}$)      & 0.84 & 1.00   & 1.00   \\
Accuracy ($\pm 10^{\circ}$)     & 0.96 & 1.00   & 1.00  \\
\hline
\end{tabular}
\vspace{-0.2cm}
\end{center}
\label{tab:orien_real}
\end{table}
\renewcommand{\arraystretch}{1}

In this section, we present our experiments on real-world outdoor AR scenarios. In this section, we first describe our system configuration and implementation details Sec.~\ref{ar_system_details}. Then, we discuss the details of our collected navigation sequences in Sec.~\ref{datasets2} and experimental results on these sequences in Sec.~\ref{ar_results}. The experiments show the impact of cross-view geo-localization based cold-start on overall navigation performance. We also present detailed 360-degree orientation estimation performance on these sequences. Finally, we show examples of AR insertions in Fig. \ref{fig:ar}.


\subsubsection{System Details}
\label{ar_system_details}
As mentioned earlier, to create a smooth AR experience for the user, we need the augmented objects to be placed at the desired spot continuously and not drift over time. This can only be achieved by accurate and consistent geo-registration in real-time on a mobile device. To achieve this, our AR system consists of a helmet-mounted sensor platform, a compact computer mounted on the backpack, and a video see-through head-mounted display (HMD). Specifically, our framework is currently implemented and executed on an MSI VR backpack computer (with Intel Core i7 CPU, 16 GB of RAM, and Nvidia GTX 1070 GPU). Our AR renderer is Unity3D based real-time renderer, which can also handle occlusions by real objects when depth maps are provided. The rendering is done on a GOOVIS HMD~\footnote{https://goovislife.com/collections/head-mounted-display} on the helmet.
Our sensor package includes Intel Realsense D435i~\footnote{https://www.intelrealsense.com/depth-camera-d435i/} and a GPS device. Intel Realsense is our primary sensor, which contains a stereo camera, RGB camera, and IMU. The computation of EKF-based visual-inertial odometry takes about 30 msecs on average for each video frame. The cross-view geo-registration process (with neural network feature extraction and reference image processing) takes an average of 200 msecs to process an input image. In the geo-registration module, we utilize the Neural Network model trained on the CVUSA\cite{zhai2017predicting} dataset.

Our reference satellite data is in geographic coordinates (latitude, longitude). We use a digital elevation model (DEM) to obtain the height of the terrain. The inserted virtual objects are placed by the renderer picking latitude and longitude coordinates using Google Earth and corresponding height using the same terrain model.


\subsubsection{Collected Navigation Sequences} 
\label{datasets2}
We collect 3 sets of navigation sequences by walking around in different places across United States. Our data was captured at 15 Hz. For these test sequences, differential GPS and magnetometer devices are used as additional sensors to create ground-truth poses for evaluation. Note, we do not use these additional sensors in our outdoor AR system to generate results. The ground camera RGB images are of 69 degrees horizontal FoV (a color camera from Intel Realsense D435i). For all the datasets, we have corresponding geo-referenced satellite imagery for the region collected from USGS EarthExplorer. We also collect Digital Elevation Model data from USGS, which we use to estimate the height.



Our first set is collected in a semi-urban location in Mercer County, New Jersey.  This set comprises three sequences with a total duration of 32 minutes and a trajectory/path length of 2.6 km. All these sequences cover both urban and suburban areas. The collection areas have some similarities to the benchmark datasets (e.g., CVUSA) in terms of the number of distinct structures and a combination of buildings and vegetation. The second set of sequences was collected in Prince William County, Virginia. This set comprises of two sequences with a total duration of 24 minutes and a trajectory length of 1.9 km. One of the sequences is collected in an urban area and the other is collected in a golf course green field. The sequence collected while walking on a green field is especially challenging as there are minimal man-made structures (e.g., buildings, roads) in the scene. The third set is collected in Johnson County, Indiana. This set comprises two sequences with a total duration of 14 minutes and a trajectory length of 1.1 km. These sequences are collected in a rural community with few man-made structures.

\subsubsection{Geo-Registration Results}
\label{ar_results}
\noindent \textbf{Continuous Global Orientation Estimation Performance:} We perform full $360$ degree heading estimation on the real-world navigation sequences described in Sec.~\ref{datasets2} and the results are reported in Table.~\ref{tab:orien_real}. In these experiments, we accumulate predictions over a sequence of frames for $10$ seconds based on Algo.~\ref{algo:seq}. We report accuracy values when the differences between the predicted heading and its ground-truth heading are within $\pm 2^{\circ}$, $\pm 5^{\circ}$, and $\pm 10^{\circ}$. We also report accuracy values with respect to different FoV coverage.

\begin{figure}[]
    \centering \includegraphics[width=0.48\textwidth]{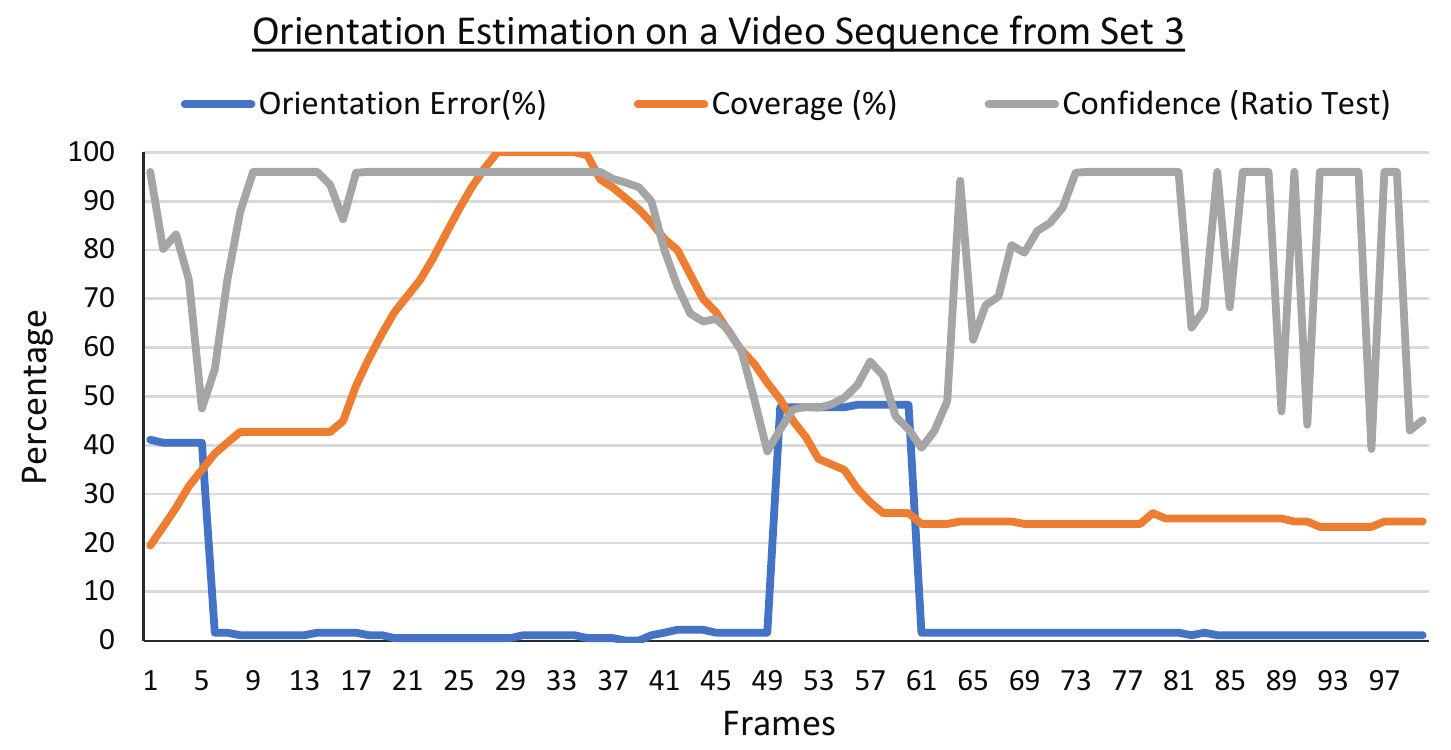}
  \caption{Orientation Error, FoV Coverage and Ratio-test score on a short sequence of frames from Set 3. We provide the related video sequence showing Ground-Truth and predicted orientation estimates matching to aerial reference as additional material 1.}
  \label{fig:plot}%
\end{figure}

\begin{figure*}[t]
    \centering
    \includegraphics[width=0.96\textwidth]{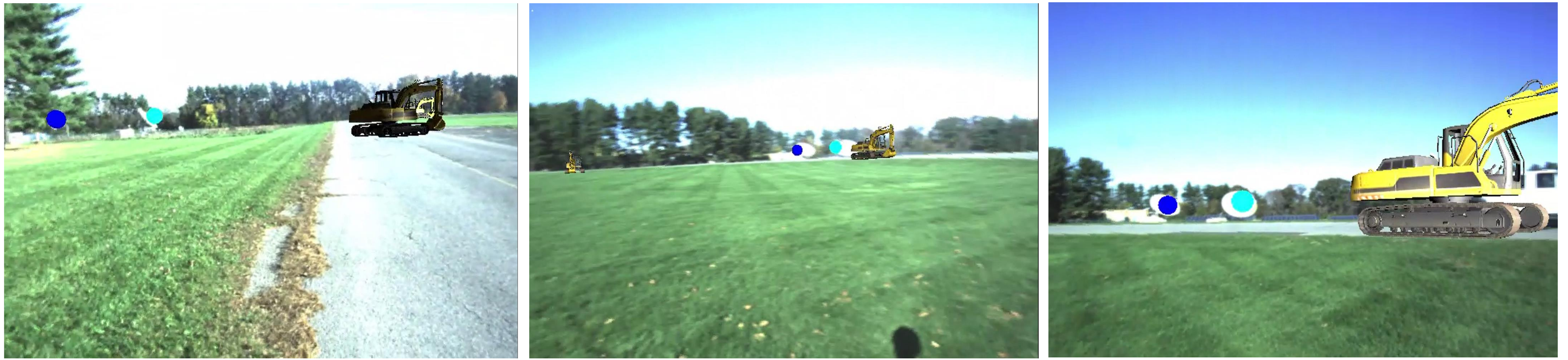}
    \vspace{-0.2cm}
  \caption{\textbf{Example screenshots from a live outdoor Augmented Reality experiment.} These images depict a semi-urban scene with two satellite dishes marked with blue and cyan spheres. The scene also has two synthetic excavators inserted in the Augmented Reality display. The three frames are extracted from one continuous walk done by a person in Mercer County, NJ. Each of these frames is taken from different perspectives but the inserted objects appear at the correct spot. The second image also has some motion blur, yet our pipeline is able to perform reliable AR insertions. Please refer to the supplementary video for the full AR sequence.}
  \label{fig:ar}%
\end{figure*}

We observe that our global heading estimation approach is able to provide reliable orientation estimates in most cases. We find huge improvement utilizing the sequence of frames and temporal information from the Nav. system, compared to using a single frame independently. For example, we find the average accuracy($\pm 2^{\circ}$) using a single frame independently is $21\%$, compared to the average accuracy of $58\%$ with our approach. We find from Table.~\ref{tab:orien_real} that orientation estimation performance using our approach is consistent across datasets. We observe the accuracy in Set 2 is slightly lower because part of the set was collected in an open green field. This is significantly different from our training set and the view has limited context nearby for matching. We observe the best performance in Set 3 even though it was collected in a rural area. We believe this is because this set was mostly recorded by walking along streets. The videos also has several portions where we look around, this helps our NN model and matching algorithm to generate high confidence estimates. We also observe from our results that with an increase in FoV coverage, the heading estimation accuracy increases as expected. 

In Fig.~\ref{fig:plot}, we plot orientation (i.e., heading) error, FoV Coverage, and confidence score based on ratio test over a frame sequence from Set 3. All the values are reported in percentages (e.g., a heading error of 50$\%$ concurs with $180^{\circ}$ error). It is evident from the plot that as coverage increases, heading error decreases. We also observe that in the middle of the sequence, our approach estimates incorrect heading (of about $180^{\circ}$) for a few frames. Such a large error happens in most cases when the reference aerial image crop is symmetric with respect to the image center. For example, there might be a scene where a road occupies the middle of the scene with green fields on both sides. This leads to high ambiguity and uncertainty in estimation. Encouragingly, we can use the ratio test score and FoV coverage of frame sequence to avoid these false detections. We provide supplementary video showing query frame, satellite crop, GT, and estimated heading for this sequence.

\renewcommand{\arraystretch}{1.4}
\begin{table}[t]
\caption{Overall Navigation Performance - Error (in meters) compared to Differential GPS Tracklets.}
\centering
    \resizebox{1\linewidth}{!}{
\begin{tabular}{p{3.5cm}ccc}
\hline
\multicolumn{1}{l|}{Systems}                                    & \multicolumn{1}{p{1.2cm}|}{RMS ~~ Error} & \multicolumn{1}{p{1.2cm}|}{Median Error} & \multicolumn{1}{p{1.3cm}}{ 90th ~~~~ Percentile} \\ \hline
\multicolumn{4}{b{8cm}}{{ ~ ~ ~ ~\underline{GPS and Magnetometer available for the whole sequence.}}}                                                                             \\ \hline
\multicolumn{1}{p{3.5cm}|}{Navigation System}                              & \multicolumn{1}{c|}{2.15}         & \multicolumn{1}{c|}{1.59}            & \multicolumn{1}{c}{3.11}             \\ \hline
\multicolumn{4}{b{8cm}}{{ ~ \underline{GPS available for the whole sequence. Magnetometer not available.}}}                                                                 \\ \hline
\multicolumn{1}{p{3.5cm}|}{Navigation System + Cross-View Geo-Reg. Module} & \multicolumn{1}{c|}{2.32}         & \multicolumn{1}{c|}{1.66}            & 3.64            \\ \hline
\multicolumn{4}{b{8.3cm}}{{ \ul GPS  Challenged Case (Only an Initial location estimate available).}} \\
\multicolumn{4}{b{8.3cm}}{{ ~~~~~~~~~~~~~~~~~~~~~~~~~~~~~~ \ul Magnetometer not available.}} \\ \hline
\multicolumn{1}{p{3.5cm}|}{Navigation System + Cross-View Geo-Reg. Module} & \multicolumn{1}{c|}{2.51}         & \multicolumn{1}{c|}{1.89}            & 3.79            \\ \hline
\multicolumn{4}{b{8cm}}{{ ~ ~ ~ ~\underline{GPS and Magnetometer available for the whole sequence.}}} \\
\multicolumn{4}{b{8.3cm}}{{ ~~~~~~~~~~~ \ul Cross-View Geo-Registration Module is also used.}} \\ \hline
\multicolumn{1}{p{3.5cm}|}{Navigation System + Cross-View Geo-Reg. Module} & \multicolumn{1}{c|}{2.08}         & \multicolumn{1}{c|}{1.48}            & 3.08          \\ \hline
\end{tabular}}
\label{tab:nav}
\end{table}
\renewcommand{\arraystretch}{1}

\vspace{0.1cm}
\noindent \textbf{Overall Navigation Performance with Cold-Start:} We present the overall navigation performance when cold-start is performed with our geo-registration module. We compare the tracklets with Differential GPS tracks to calculate these results. We mainly consider three cases. First, GPS and magnetometer devices are available in the sensor package and working well. In such a case, we do not apply our cross-view geo-registration. Second, GPS is available for the whole sequence, but a magnetometer is not available. In this case, we apply our cross-view geo-registration module for cold-start global heading estimation. Third, we consider a GPS-challenged scenario where an initial rough location estimate is available. The magnetometer is not available. In this case, we apply our cross-view geo-registration module for cold-start global heading and position estimation. From Table~\ref{tab:nav}, we observe when GPS and magnetometer are both available and working well, the best performance is achieved. We also observe that Cold-start with our geo-registration module helps the navigation system to still achieve reliable navigation performance even when GPS is challenged and a magnetometer is not available for heading estimation. In Table~\ref{tab:nav}, we primarily focused on evaluating whether cold-start with our geo-registration module helps the navigation system to achieve comparable performance in case the magnetometer and/or GPS are not working well. For completeness, we also evaluate a fourth case when the Geo-Registration module is used for cold-start together with GPS and magnetometer available and working well. We find a slight improvement in performance compared to the first case. Please note that the magnetometer is not available in our sensor package for the live system (Section 5.2.1). We have only used magnetometer and differential GPS as additional sensors to create ground-truth poses for evaluation (Section 5.2.2).

\begin{figure*}[]
    \centering \includegraphics[width=0.92\textwidth, height=0.27\textwidth]{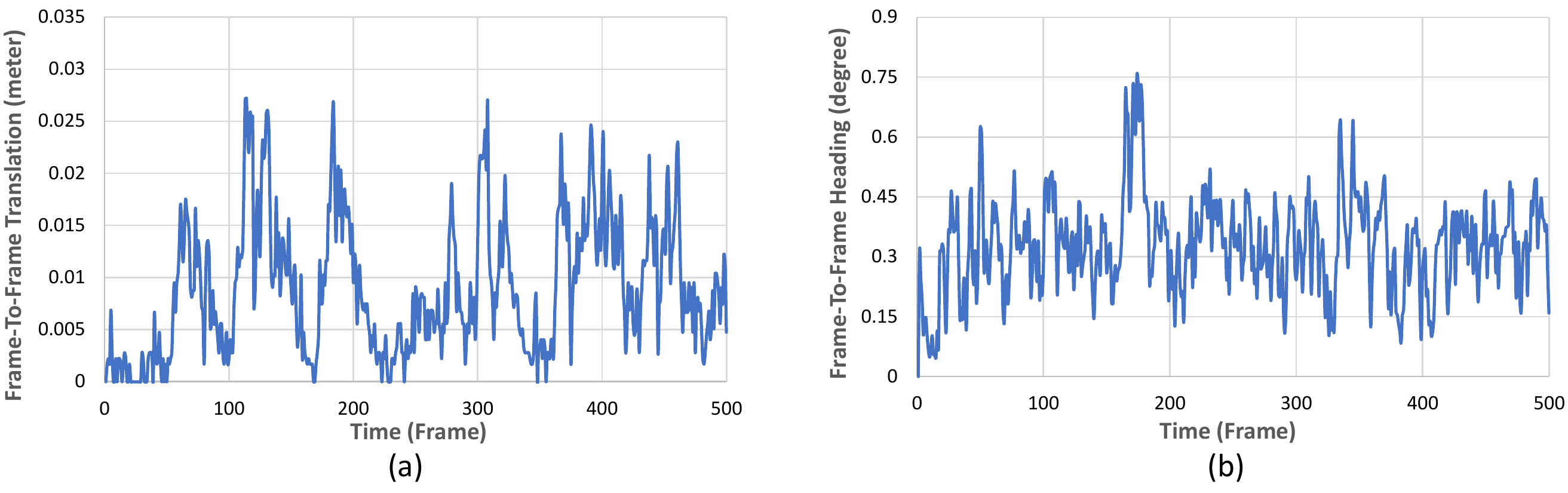}
    \vspace{-0.3cm}
  \caption{(a) Frame-to-frame estimated translation computed by our system and (b) Frame-to-frame estimated heading change computed by our system over a 500-frame sequence from Mercer County Set.}
 \vspace{-0.3cm} 
  \label{fig:f2f}%
\end{figure*}

We present a few augmented reality insertion examples from Set1 in Fig.~\ref{fig:ar} when cold-start is performed with our cross-view geo-registration module. We insert virtual objects at different locations (picked using Google Earth) based on the refined pose estimated by our navigation system. We record the augmented reality insertion video, which is seen by the user from the VST HMD. We observe that the positions of the inserted virtual objects are consistent in these snapshots in Fig.~\ref{fig:ar}, which indicates that our system is able to provide consistent pose estimation for a long period. The consistency of augmented content across consecutive frames can be seen qualitatively in the supplementary AR insertion video. To quantitatively evaluate the reliability of estimated pose across frames, we show frame-to-frame pose translation estimated by our system in Fig.~\ref{fig:f2f} (a). Here, we show the translation over a 500-frame period taken from the Mercer County Set1. The translation between the frames is expected to be very smooth as the users' walking speed is unlikely to change significantly in the short time period (such as one frame, 0.0333 seconds). From Fig.~\ref{fig:f2f} (a), we observe that the frame-to-frame translation is smooth and there are no sudden large peaks in the curve. In Fig.~\ref{fig:f2f} (b), we show frame-to-frame heading change over the same frame period and observe that the change between consecutive frames is also smooth. We observe a similar trend across all the navigation sequences. Based on the results and insertion videos, we find the augmented contents to be very consistent with minimal jitter across consecutive frames.


\section{Limitations and Discussion}
\label{future_work}
Our work presents promising results and provides precise geo-registration estimates in most cases for a smooth outdoor AR experience.  However, there are still some limitations. 

First, similar to most machine learning models, our model also suffers from cross-dataset generalization issues. The neural network models are trained using benchmark datasets that include pairs of street-view images and reference image crops (covering a small area centering camera location). The ground images in the benchmark dataset are mostly collected in urban and semi-urban settings on streets. For these environments, our trained model works well. The ground camera typically covers a scene that is short or mid-range distance (tens or hundreds of meters) till it gets obstructed by a building or vegetation. For such street-view ground queries, the fixed-size satellite image crops are able to provide sufficient context for cross-view matching. Our system is able to utilize these nearby contexts to provides reliable estimates of geo-poses. 

However, in some environments (e.g., deserts), most meaningful objects (buildings, trees, mountains) in the ground viewpoint can be several hundreds of meters away (outside the coverage area of a reference aerial image crop). To overcome this limitation, we believe utilizing semantics of the scene and multiple spatial-scale feature maps from satellite images covering a larger area would help in significantly improving the reliability of the system. 

Second, vision-based geo-localization systems may also suffer significantly due to perceptual changes caused by changes in weather, seasons, and time of day. Based on prior works (e.g.~\cite{toft2020long}), we believe our system will be less susceptible to these changes due to relying on sequential information from multiple frames for prediction, rather than predicting with a single frame. Moreover, we observe orientation estimation relies more on man-made structures and road boundaries, which are mostly season invariant. In the future, we plan to investigate further the impact of environmental changes on our system by explicitly collecting datasets at different times of the day and seasons.

Third, our approach currently estimates heading at 1-degree granularity. For long-range outdoor AR applications, it might be desirable to achieve sub-degree orientation estimates. One approach to address this issue would be to adopt a parabola-fitting sub-bin refinement approach (e.g., similar to subpixel refinement in stereo matching \cite{nehab2005improved}) to estimate heading at sub-degree precision. These can be future directions of our research.



\section{Conclusion}
\label{conclusion}

We propose the first transformer neural network-based cross-view geo-localization method, which jointly predicts the location and orientation of ground image queries. Our method achieves state-of-the-art results on benchmark datasets. The proposed geo-localization approach enables practical outdoor Augmented Reality applications. We develop a navigation system that utilizes the cross-view geo-localization module to find the geo-pose of the camera. Experiments on several navigation sequences demonstrate the robustness and applicability of our proposed system for outdoor AR applications.  We hope this contribution will help the community further its mission towards a smooth and realistic AR experience.

\acknowledgments{The work for this program was conducted with the support of the Office
of Naval Research (ONR) Contracts N00014-19-C-2025. We thank
ONR for their support. The views, opinions, and findings contained in
this report are those of the authors and should not be construed as an
official position, policy, or decision of the United States government.}

\bibliographystyle{abbrv-doi}

\bibliography{template}
\end{document}